\documentclass[10pt,twocolumn,letterpaper]{article}

\usepackage[pagenumbers]{cvpr} 

\definecolor{cvprblue}{rgb}{0.21,0.49,0.74}
\usepackage[pagebackref,breaklinks,colorlinks,allcolors=cvprblue]{hyperref}
\usepackage{mathtools}
\usepackage{bm}       
\usepackage{enumitem}
\setlist[itemize]{leftmargin=1.5em, topsep=0.2em, itemsep=0.2em}
\usepackage{graphicx}     
\usepackage[table]{xcolor}
\usepackage{pgfplots}
\usepackage{pgfplotstable}
\pgfplotsset{compat=1.18} %
\usepgfplotslibrary{groupplots}
\usepackage{subcaption}
\usepackage{array}
\usepackage{multirow}
\usepackage{makecell}
\usepackage{booktabs}   
\usepackage{siunitx}
\usepackage{float}
\usepackage[accsupp]{axessibility} 
\def\confName{CVPR}
\def\confYear{2026}

\usepackage[dvipsnames]{xcolor} 
\usepackage{pifont,xcolor}

\newcommand{\badmark}{\textcolor{red}{\ding{51}}}

\title{HieraMamba: Video Temporal Grounding via\\ Hierarchical Anchor-Mamba Pooling}

\author{
Joungbin An \qquad Kristen Grauman \\
The University of Texas at Austin 
}
\begin{document}
\maketitle

\begin{abstract}
Video temporal grounding, the task of localizing the start and end times of a natural language query in untrimmed video, requires capturing both global context and fine-grained temporal detail. This challenge is particularly pronounced in long videos, where existing methods often compromise temporal fidelity by over-downsampling or using fixed windows. We present HieraMamba, a hierarchical architecture that preserves temporal structure and semantic richness across scales. At its core are Anchor-MambaPooling (AMP) blocks, which utilize Mamba’s selective scanning to produce compact anchor tokens summarizing video content across scales. We further introduce anchor-conditioned and segment-pooled contrastive losses---two complementary objectives that encourage anchors to retain local detail while remaining globally discriminative. HieraMamba sets a new state-of-the-art on Ego4D-NLQ, MAD, and TACoS, demonstrating precise, temporally faithful localization in long, untrimmed videos.\footnote{Project webpage: \url{https://vision.cs.utexas.edu/projects/hieramamba}.}
\end{abstract}
\vspace{-6mm}
\section{Introduction} \label{sec:intro}
\begin{figure}[!t]
  \centering
  \includegraphics[width=1.0\linewidth]{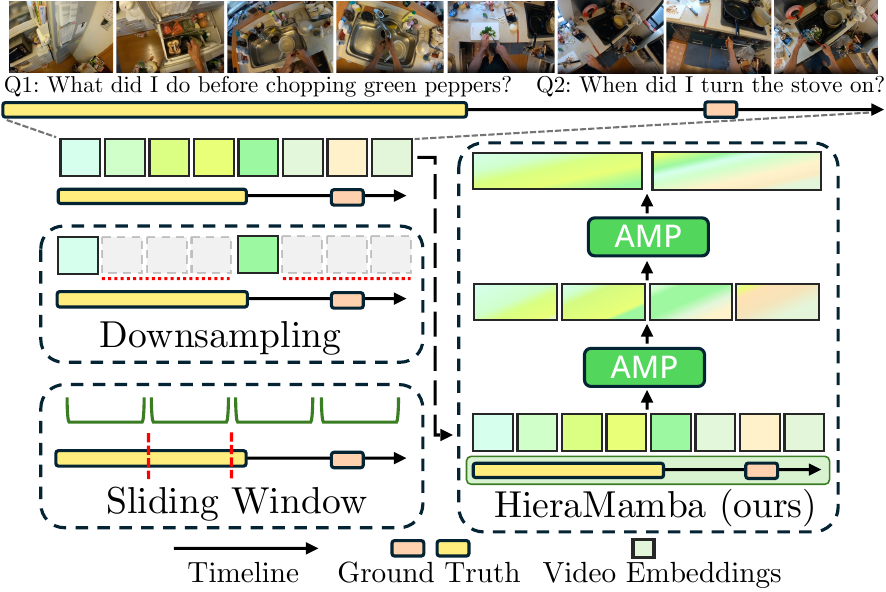}
\caption{
\textbf{HieraMamba} enables \emph{hierarchical}, \emph{linear-time} temporal grounding in long untrimmed videos.  
\emph{Top row:} 
a cooking clip and two queries—Q1's answer spans a long interval, Q2's a short one.  
\emph{Middle left:} uniform downsampling (gray squares) drops frames and loses evidence for the queries.  
\emph{Bottom left:} fixed sliding windows split segments at window boundaries but risk fragmenting (red dashed lines).  
\emph{Right:} 
HieraMamba stacks Anchor-MambaPooling (AMP) blocks to build a multi-scale temporal hierarchy that, unlike traditional feature pyramids, uses Mamba to compress and refine information at each level, enabling efficient fine-to-coarse abstraction and precise, query-specific localization.
For example, the brief `stove on' moment in Q2 is captured by fine-scale embeddings in the first layer, 
while Q1's broader `prepping ingredients' context is naturally represented by the coarser embeddings at the top layer. 
} 
  \label{fig:teaser}
  \vspace{-3mm}
\end{figure}

Humans readily access broad episodic memories, answering `What did I do this morning?' with ease, while simultaneously being able to pinpoint fine-grained details like `Where did I leave my keys?' or `Did I lock the front door?' Our memory naturally navigates across multiple temporal scales, shifting seamlessly from the overall layout of a room to the precise motion of our fingertips: an inherently hierarchical process~\cite{Kolibius2023Episodic,Mishkin1997Hierarchy}.  

Video temporal grounding, the task of 
identifying the precise moment in an untrimmed video that corresponds to a language query, seeks to give machines this same `instant recall' ability. Evolving from localizing predefined actions~\cite{zeng2019gcn, zhang2022actionformer, cheng2022tallformer, xu2020gtad} to handling free-form language queries~\cite{soldan2022mad, grauman2022ego4d, tacos, zhang2020span, snag, lei2021detecting}, temporal grounding supports visual question answering across egocentric~\cite{grauman2022ego4d}, movies~\cite{soldan2022mad}, or third-person instructional videos~\cite{tacos}.

Yet replicating this multi-scale recall over continuous video remains challenging: current models struggle to faithfully preserve both the broad temporal layout and pinpoint precise moments, like when keys hit the countertop.
To close this gap, we require systems that can mirror and even surpass our hierarchical, multi-scale memory, supporting precise retrieval of desired episodes from long videos.

While many methods excel on short clips, long-form videos pose two intertwined challenges. 
\textbf{First}, minutes-to hours-long videos demand models capable of preserving temporal structure across extended sequences. 
However, many existing methods compress temporal resolution through fixed-length pooling and/or naive downsampling—either jointly~\cite{zhang2020learning, zhang2020span, soldan2021vlg, lei2021detecting} or solely through naive downsampling~\cite{snag, feng2025OSGNet, lu2025decafnet}—discarding critical cues in long videos. Others rely on fixed-window heuristics~\cite{hou2022cone, hannan2024rgnet}, which often fragment temporal structure. While these strategies help reduce the computational cost of processing long videos, they fall short in capturing dependencies that extend across long temporal spans.

\textbf{Second}, queries demand flexible temporal granularity: some require broad contextual understanding 
(e.g., what did the detective do in the library?), while others depend on subtle fine-grained motions (e.g., when did the detective pull the hidden note from the shelf?),
and many hinge on both. Traditional single-resolution methods struggle to meet these demands, often sacrificing one type of detail for the other. 

These challenges call for models that preserve temporal fidelity across scales while remaining efficient. To this end, we introduce HieraMamba, the first hierarchical state-space video model for temporal grounding in hour-long videos. Identifying quadratic self-attention as the root cause behind the downsampling and windowing heuristics of prior methods, we design a model that operates in linear time via Mamba's selective scanning, enabling long-range reasoning without sacrificing temporal resolution. We further extend this architecture to produce hierarchical embeddings, from fine to coarse, 
that capture both global context and fine-grained temporal details for precise query localization.

At its core are our novel Anchor-MambaPooling (AMP) blocks that summarize short video segments into compact anchor tokens. Each AMP block fuses these anchors with local video features through Mamba’s selective scanning, yielding both updated fine-grained features and coarse, semantically meaningful summaries. Stacking AMP blocks forms a multi-scale temporal hierarchy—analogous to a feature pyramid, but crucially, learned through token-level compression rather than naive downsampling~\cite{snag, feng2025OSGNet, lu2025decafnet, zhang2022actionformer}.
This novel design preserves temporal detail across scales while remaining linear in cost, enabling precise grounding even in hour-long footage.
See Figure~\ref{fig:teaser}.

To enrich the AMP embeddings and preserve both global semantics and local details, we introduce two complementary contrastive objectives: 
an anchor-conditioned contrastive (ACC) loss, which encourages each generated anchor to better compress and represent the frames it summarizes while remaining distinct from unrelated ones, and a segment-pooled contrastive (SPC) loss, which pools each ground-truth segment into a single anchor and contrasts it against non–ground-truth moments, thereby enhancing the discriminability of localized event representations.
Together, ACC and SPC refine the hierarchical embeddings into compact, semantically discriminative representations for precise grounding even on hour-long videos.

We evaluate HieraMamba on three long-video temporal grounding benchmarks, Ego4D-NLQ~\cite{grauman2022ego4d}, MAD~\cite{soldan2022mad}, and TACoS~\cite{tacos}, where it consistently outperforms prior methods. These results validate the effectiveness of our hierarchical architecture and contrastive learning framework in preserving temporal fidelity and achieving precise grounding, while retaining the linear-time scalability of Mamba.

In summary, we (i) introduce HieraMamba that utilizes our novel AMP block for hierarchical compression, (ii)  propose two contrastive objectives to enhance semantic precision, and (iii) achieve state-of-the-art grounding performance on Ego4D-NLQ, MAD, and TACoS, validating the effectiveness of our model design and learning objectives.

\section{Related Work} \label{sec:relatedworks}

\paragraph{State‐Space Models in Image and Video Understanding.}
State‐space models (SSMs) have emerged as a compelling alternative to Transformers and RNNs for long‐range sequence modeling. Foundational works, HiPPO~\cite{gu2020hippo,gu2021combining} and S4~\cite{s4}, show that structured state matrices can summarize past information with linear complexity. Several recent advances further improve long-term dependency modeling: Mamba adds an input-dependent state-space layer~\cite{mamba}, Mamba-2 unifies SSM and Transformer attention~\cite{mamba2}, and Hydra adds bidirectional modeling~\cite{hydra}.

Originally developed for non-visual sequential data such as text and audio, these architectures have been adapted for image and video understanding, including as backbones and downstream modules 
for spatial context~\cite{liu2024vmamba, ma2025tinyvim, hatamizadeh2025mambavision}, spatio-temporal graphs~\cite{chaudhuri2024simba}, state-space updates on raw frames~\cite{li2024videomamba,park2024videomamba,lu2024snakes}, hybrid SSM-Transformer architectures~\cite{islam2022long,islam2023efficient, heo2025autoregressive}, and token-efficient compression for VideoQA~\cite{islam2025bimba,jiang2025token}.
Unlike these end-to-end, frame-level variants, our Anchor-MambaPooling operates on clip embeddings extracted from off-the-shelf video backbones (e.g., TimeSformer~\cite{bertasius2021space}, InternVideo~\cite{wang2022internvideo}). 
Inserting lightweight blocks on pre-computed embeddings decouples spatial and temporal modeling while compressing video features into efficient multi-scale temporal representations.

\vspace{-3mm}
\paragraph{Video Temporal Grounding.}
With applications in personal assistants, human-robot interaction, and video editing, video temporal grounding has evolved rapidly.
Earlier methods focus on short clips (less than a minute), establishing the grounding as either candidate proposal ranking~\cite{anne2017localizing,gao2017tall,zhang2020learning,yuan2019semantic,wang2021structured,soldan2021vlg,chen2023joint} 
or direct boundary regression~\cite{ghosh2019excl,zhang2020span,zeng2020dense,mun2020LGI}.
While effective on minute-scale clips, these methods struggle to handle
several-minute or hour-long videos due to design choices that limit long-range reasoning—most notably the quadratic cost of self-attention.

In \emph{Long-Video Temporal Grounding} (LVTG), initial approaches constrain the sequence length through fixed-length pooling or truncation~\cite{zhang2020learning, zhang2020span, soldan2021vlg, lei2021detecting, ramakrishnan2023spotem}, which reduces computation but discards fine temporal detail.
Subsequent approaches~\cite{hou2022cone,hannan2024rgnet} preserve more context using fixed-size sliding windows, though window boundaries often disrupt temporal coherence. Recent efforts~\cite{snag,lu2025decafnet,feng2025OSGNet} introduce multi-scale modeling through windowed attention~\cite{zhang2022actionformer}, yet those multiple scales still arise from uniform downsampling or coarse pooling. 
In contrast, HieraMamba processes full-length video sequences without the compromises of prior approaches, effectively harnessing the linear-time efficiency of state-space models to achieve scalable and fine-grained temporal grounding.
Table~\ref{tab:method_characteristics} summarizes these trade-offs.

\vspace{-1mm}
\paragraph{Hierarchical Video Understanding.}
A complementary line of work organizes long videos hierarchically to manage scale and structure. Ego4D Goal-Step~\cite{song2023ego4d} decomposes procedures from goals to steps, VideoReCap~\cite{islam2024videorecap} performs recursive long-video captioning, VideoTree~\cite{wang2025videotree} builds adaptive tree-based representations for LLM reasoning, and OpenHOUSE~\cite{kang2025open} structures narratives across coarse-to-fine timescales. In grounding and action localization, hierarchy is often implemented as temporal feature pyramids. ActionFormer~\cite{zhang2022actionformer} introduced multiscale representations built by strided pooling over time, and follow-ups such as SnAG~\cite{snag}, DeCafNet~\cite{lu2025decafnet}, and OSGNet~\cite{feng2025OSGNet} refine the design but still rely on fixed windows or uniform downsampling. 
While sharing the same hierarchical intuition, HieraMamba differs fundamentally in mechanism. It builds multi-scale representations via Anchor-MambaPooling, which compresses clip embeddings into hierarchical anchor tokens and optimizes them with anchor-conditioned and segment-pooled contrastive objectives. Together, these components preserve local detail and global context, enabling long-range reasoning and precise localization across multiple temporal scales.

\begin{figure*}[t]  
  \centering
  \includegraphics[width=\textwidth]{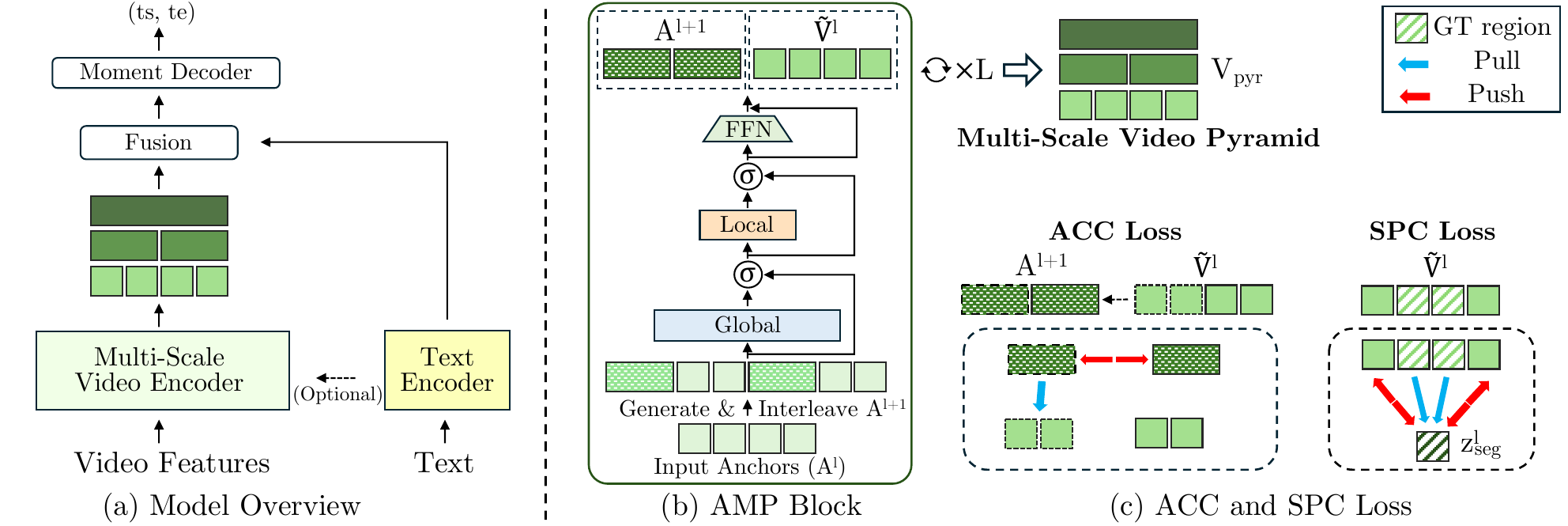}
    \caption{
        \textbf{Overview of the HieraMamba Architecture.}
        \textbf{(a)} Frozen backbones extract video clip and text token features. The hierarchical video encoder, a stack of $L$ AMP blocks, builds a multi-scale pyramid $\mathcal{V}_{\text{pyr}}$, which is fused with text features and decoded to predict timestamps. 
        \textbf{(b)} Each AMP block receives anchors from the previous layer ($A^{(l)}$), interleaves them with new compressed anchors ($A^{(l+1)}$), applies a bidirectional Mamba scan for global context, and refines local details. The block outputs refined tokens ($\tilde V^{(l)}$) and downsampled anchors ($A^{(l+1)}$) fed to the next block. Repeating this $L$ times and collecting the refined outputs $\{\tilde V^{(l)}\}_{l=0}^{L-1}$ forms the multi-scale hierarchy $\mathcal{V}_{\text{pyr}}$. 
        \textbf{(c)} Two contrastive losses guide training. The self-supervised \textbf{ACC} loss enforces hierarchy consistency by pulling anchors toward their constituent frames and pushing from distant anchors. The supervised \textbf{SPC} loss provides semantic alignment between ground-truth segments and surrounding context. Together, they yield compact, distinctive, and query-aligned anchors.
    }
  \label{fig:main}
  \vspace{-4mm}
\end{figure*}

\section{Preliminaries} \label{sec:preliminaries}
To motivate our design, we first summarize the state-space formulations that enable efficient long-range dependency modeling in sequential data.

\subsection{State Space Models (SSMs)}  
State Space Models (SSMs) model sequential data using latent dynamics governed by linear systems. The continuous-time formulation~\cite{gu2020hippo} is:
\vspace{-1mm}
\begin{equation}
\frac{d\boldsymbol{h}(t)}{dt} = \mathbf{A}\boldsymbol{h}(t) + \mathbf{B}\boldsymbol{x}(t), \quad
\boldsymbol{y}(t) = \mathbf{C}\boldsymbol{h}(t) + \mathbf{D}\boldsymbol{x}(t),
\vspace{-1mm}
\end{equation}
where $\boldsymbol{x}(t)$ is the input, $\boldsymbol{h}(t)$ is the latent state, and $\boldsymbol{y}(t)$ is the output. Discretization (e.g., via zero-order hold) yields~\cite{s4}:
\vspace{-1.5mm}
\begin{equation}
\boldsymbol{h}_k = \overline{\mathbf{A}} \boldsymbol{h}_{k-1} + \overline{\mathbf{B}} \boldsymbol{x}_k, \quad
\boldsymbol{y}_k = \overline{\mathbf{C}} \boldsymbol{h}_k + \overline{\mathbf{D}} \boldsymbol{x}_k,
\vspace{-1mm}
\end{equation}
where $\overline{\mathbf{A}}, \overline{\mathbf{B}}, \overline{\mathbf{C}}, \overline{\mathbf{D}}$ are the learned, fixed transition and projection matrices. Classical SSMs allow for efficient linear-time inference, but the fixed nature of these parameters limits flexibility and expressiveness.

\subsection{Mamba: Selective State Space Models}
Mamba~\cite{mamba} introduces a data-dependent SSM layer with token-wise, input-conditioned parameters. Specifically, for each input $\boldsymbol{x}_k$, it computes dynamic modulation terms $\mathbf{B}_k$, $\mathbf{C}_k$, and step size $\Delta_k$, and updates the scan state as:
\vspace{-1mm}
\begin{equation}
\tilde{\boldsymbol{y}}_k = \Delta_k \cdot \left( \mathbf{A} \tilde{\boldsymbol{y}}_{k-1} + \mathbf{B}_k \odot \boldsymbol{x}_k \right), \quad
\boldsymbol{y}_k = \mathbf{C}_k \cdot \tilde{\boldsymbol{y}}_k.
\vspace{-1mm}
\end{equation}
This formulation allows Mamba to selectively modulate its state based on input content, combining the long-range modeling benefits of SSMs with the adaptability of attention, while retaining linear-time inference. Mamba-2~\cite{mamba2} further establishes a structured duality between attention and SSMs, showing that attention weights can be emulated via state-space filters with appropriate kernelization.

This selective, token-aware structure makes Mamba 
well-suited for long video sequences, where modeling both local and global dependencies efficiently is crucial. In our work, we incorporate these properties into the proposed \textit{Anchor-MambaPooling} block, which hierarchically compresses video features into compact, semantically meaningful anchors using Mamba’s linear-time state-space scans.

\begin{table}[t]
\centering
\footnotesize
\setlength{\tabcolsep}{5pt}
\renewcommand{\arraystretch}{1.12}

\resizebox{\columnwidth}{!}{
\begin{tabular}{lccccc}
\toprule
\textbf{Method} & \shortstack{\textbf{Naive} \\ \textbf{Downsampling}} & 
\shortstack{\textbf{Fixed-Length} \\ \textbf{Pooling}} & 
\shortstack{\textbf{Quadratic} \\ \textbf{Cost}} & 
\shortstack{\textbf{Sliding} \\ \textbf{Window}} & 
\shortstack{\textbf{Ego4D} \\ \textbf{Avg.R$\uparrow$}} \\
\midrule
2D-TAN~\cite{zhang2020learning}      & \badmark & \badmark & \badmark & —          & 6.46 \\
VSLNet~\cite{zhang2020span}      & \badmark & \badmark & \badmark & —          & 12.49 \\
M-DETR~\cite{lei2021detecting} & \badmark & \badmark & \badmark & \badmark   & 12.46 \\
CONE~\cite{hou2022cone}        & —        & —        & \badmark & \badmark   & 17.67 \\
RGNet~\cite{hannan2024rgnet}       & —        & —        & \badmark & \badmark   & 21.81 \\
SnAG~\cite{snag}        & \badmark & —        & —        & —          & 23.08 \\
DeCafNet~\cite{lu2025decafnet}    & \badmark & —        & —        & —          & 24.44 \\
OSGNet~\cite{feng2025OSGNet}      & \badmark & —        & —        & —          & 22.46 \\
\textbf{HieraMamba (ours)} & —      & —        & —        & —          & \textbf{25.66} \\
\bottomrule
\end{tabular}}
\vspace{-2mm}
\caption{\textbf{Method characteristics and limitations.}
Red checks (\badmark) indicate undesirable properties that degrade long-video performance; ``—'' indicates the property is absent. 
By avoiding all such limitations, our method achieves the best accuracy (shown here in terms of Ego4D average recall). 
}
\label{tab:method_characteristics}
\vspace{-4mm}
\end{table}

\section{Approach} \label{sec:approach}
We first formalize the problem and then present our multi-scale architecture. After overviewing our model, we introduce the Anchor-MambaPool blocks, followed by our two training objectives.

\subsection{Long Video Temporal Grounding}
Given an untrimmed video represented by features $V = \{v_i\}_{i=1}^{L_{V}} \in \mathbb{R}^{L_{V}\times D_v}$, and a natural language query represented by word embeddings $Q = \{w_j\}_{j=1}^{L_{Q}} \in \mathbb{R}^{L_{Q}\times D_q}$, the goal is to learn a function $f(V, Q) \rightarrow (t_s, t_e)$. Here, $(t_s, t_e)$ are the predicted start and end times of the video segment that provides the answer to $Q$. We follow standard temporal grounding benchmarks, where each query's answer corresponds to a single contiguous segment within the video.

\subsection{Model Overview}
Figure~\ref{fig:main} overviews our architecture. It processes raw video frames and text to produce embeddings, which are then refined by specialized video and text encoders before being fused to predict the final timestamps.

\noindent\textbf{Feature Extraction.}
Raw frames are encoded by a frozen video backbone (e.g., EgoVLP~\cite{lin2022egocentric}) into clip-level features $V$, while the query is embedded by a frozen text model (e.g., CLIP text encoder~\cite{radford2021learning}) into $Q$. Freezing both backbones maintains the pipeline's modularity and efficiency.

\noindent\textbf{Video and Text Encoders.}
The \textbf{text encoder} uses a stack of standard transformers to refine the initial word embeddings $Q$, producing contextually enriched query embeddings $E \in \mathbb{R}^{L_Q \times D_q}$. 
The \textbf{multi-scale video encoder}, our key contribution, is a hierarchical stack of $L$ Anchor–MambaPooling (AMP) blocks (see Section~\ref{sec:amp_block}). 
This stack progressively refines the clip features $V^{(0)} = V$ and produces a multi-scale feature pyramid 
$\{\tilde{V}^{(l)}\}_{l=0}^{L-1}$ that captures temporal context at increasing granularity.

\noindent\textbf{Fusion and Decoding.}
The multi-scale video feature pyramid $\{\tilde{V}^{(l)}\}_{l=0}^{L-1}$ and text embeddings $E$ are fed into a cross-modal attention module to produce a fused representation $X_{\text{fused}}$, which is then passed to a lightweight decoder~\cite{zhang2022actionformer} that regresses the final start and end timestamps $(t_s, t_e)$.

\subsection{Anchor--MambaPooling Block}
\label{sec:amp_block}

The \emph{Anchor--MambaPooling} (AMP) block is a \textbf{stackable} module that constructs a hierarchical, multi‑scale representation of a video stream. At each level it (i) \emph{refines} features at the current temporal resolution and (ii) \emph{summarizes} them into a compact set of \emph{anchor} tokens for the next, coarser scale, harnessing Mamba's state‑space selective scan to model long‑range dependencies with linear complexity.

Let $A^{(0)} = V^{(0)}\!\in\!\mathbb{R}^{L_0\times D_v}$ denote the initial backbone features fed into the first AMP block. \textbf{Layer~0} outputs (i) a refined sequence $\tilde V^{(0)}\!\in\!\mathbb{R}^{L_0\times D_v}$ and (ii) an anchor set $A^{(1)}\!\in\!\mathbb{R}^{L_1\times D_v}$, where each anchor represents a compact summary of its local temporal window, and $L_1=\lceil L_0/s \rceil$ for stride $s$. For any layer $l>0$, the block receives $A^{(l)}$ and returns a refined version $\tilde V^{(l)}$ together with a further-downsampled anchor set $A^{(l+1)}$.

Repeating this process for $L$ layers yields the feature pyramid
\[
  \mathcal{V}_{\text{pyr}} = \{\tilde V^{(0)},\, \tilde V^{(1)},\, \dots,\, \tilde V^{(L-1)}\},
\]
where each $\tilde V^{(l)}$ provides context at its characteristic temporal granularity for downstream grounding.

Unlike fixed pooling or strided convolutions that compress indiscriminately, AMP performs \textbf{content‑aware abstraction}: it learns to distill salient segments into anchors that propagate up the hierarchy, producing a compact yet faithful representation that supports scalable long‑range reasoning and precise temporal localization.  Figure~\hyperref[fig:main]{\ref*{fig:main}{(b)}} visualizes the data flow.
We explain this property in \S\ref{sec:anchor-init}--\S\ref{sec:amp-details}, which detail (i) anchor generation \& interleaving, (ii) the dual global–local encoding scheme, and (iii) the gated design choices that complete the AMP block.

\subsubsection{Anchor Generation and Interleaving}
\label{sec:anchor-init}

The first stage of the AMP block is the generation and interleaving of anchor tokens. Given the first-level features $V^{(0)} \in \mathbb{R}^{L_0 \times D_v}$ and a temporal stride $s$, we instantiate one anchor every $s$ frames, yielding $A^{(1)} \in \mathbb{R}^{L_1 \times D_v}$ where $L_1$ denotes the number of anchors. Each anchor token is initialized by pooling over its local window of $s$ frames (pooling strategies evaluated in the supplementary material).

We expose these anchors and input tokens to the same selective scan by interleaving them into a single sequence
\[
  \resizebox{\linewidth}{!}{%
    $\hat V = [a_0,\,v_0,\dots,v_{s-1},\,a_1,\,v_s,\dots,v_{2s-1},\,\dots] 
      \in \mathbb{R}^{(L_0+L_1) \times D_v}$%
  },
\]
placing each anchor $a_i$ immediately \emph{before} the $s$ frames it summarizes. 
This layout preserves temporal order and enables bidirectional information flow: anchors broadcast coarse context to nearby frames, while frame-level evidence refines the anchors during the Mamba scan.

\subsubsection{Global and Local Encoding}
The anchor-interleaved sequence initially lacks the temporal cues essential for comprehensive video understanding. To address this, we enrich the representation through a combination of global and local encoding mechanisms tailored for long-form video reasoning.

Motivated by the recent success of state space models in capturing long-range dependencies, we adopt Hydra~\cite{hydra}; we find Hydra’s forward–backward scan effectively models global temporal context while preserving the linear-time complexity characteristic of Mamba.

To complement the global representation, we add a lightweight local encoder~\cite{zhang2022actionformer} focused on short-range patterns. While recent hybrid models~\cite{ren2025vamba, lieber2024jamba, hwang2025dynamic} showcase the synergy of Mamba and Transformers, our design explicitly decouples their roles: Mamba efficiently captures global structure, while a local Transformer with a narrow temporal window (e.g., size 5) provides fine-grained attention without the cost of global self-attention.

\subsubsection{Design Details of the AMP Block}
\label{sec:amp-details}
Following standard practices, each substage of the AMP block, global encoding, local encoding, and FFN, is preceded by RMS normalization~\cite{zhang2019root} followed by residual connections. Normalizations are omitted from Fig.~\ref{fig:main} for clarity.

Feature fusion between stages is controlled by a learnable sigmoid gate ($\boldsymbol{\sigma}$ in Fig.~\ref{fig:main}), providing a content-adaptive alternative to unconditional residual addition~\cite{dauphin2017language,cho2014learning,shazeer2020glu}. This gating mechanism allows the network to propagate only salient information as representations are refined up the hierarchy. The block ends with a feed-forward network that further refines the output, from which we extract (i) next-level anchor tokens $A^{(l+1)}$ summarizing salient regions for the following AMP layer and (ii) refined sequence tokens $\tilde{V}^{(l)}$ serving as current-resolution embeddings for the final pyramid or downstream decoding.

\subsection{Contrastive Objectives}
\label{sec:contrastive}

To guide the hierarchical features produced by the AMP blocks toward \emph{compact} yet \emph{discriminative} semantics, we devise two complementary losses: \textbf{anchor-conditioned contrastive} (ACC) and \textbf{segment-pooled contrastive} (SPC).

\paragraph{Anchor-Conditioned Contrastive (ACC) Loss.}
At the $l$-th layer, the AMP block produces anchors $A^{(l+1)}$ and refined sequence tokens $\tilde{V}^{(l)}$, which serve as inputs to the ACC loss.
For the hierarchical representation to be effective, anchors must meet two criteria: they should be \textbf{compact}, faithfully representing the temporal window from which they are derived, and \textbf{distinctive}, clearly separable from other anchors.
The ACC loss is a self-supervised objective applied at every layer to enforce these properties.

For \textbf{compactness}, each anchor is matched to all frame tokens within its local window using a multi-positive formulation.
For anchor $\boldsymbol{a}_i^{(l+1)}$, the positive set $\mathcal{P}_i^{(l)}$ contains all $s$ tokens
${\tilde{\boldsymbol{v}}_t^{(l)} \mid t \in [is,, is+s)}$ from $\tilde{V}^{(l)}$ belonging to its temporal span, encouraging each anchor to summarize the key information of its window at a coarser temporal scale.
To ensure \textbf{distinctiveness}, anchors are contrasted against a negative set $\mathcal{N}_i^{(l)}$ of distant anchors.
Negatives are selected with a temporal margin to avoid penalizing adjacent ones that may correspond to the same event, and their number is limited relative to positives to prevent imbalance in long videos.
This design promotes well-separated anchor representations, each capturing a distinct temporal context while remaining faithful to its local content.

Formally, after the $l$-th AMP block, we obtain $A^{(l+1)}$ and $\tilde{V}^{(l)}$, 
project both through a shared linear head, and apply a multi-positive InfoNCE loss:
\begin{equation}
\label{eq:acc_single}
\mathcal{L}_{\mathrm{acc}}\!\bigl(\boldsymbol{a}_i^{(l+1)}\bigr) =
-\log \frac{
    \displaystyle \sum_{\boldsymbol{p} \in \mathcal{P}_i^{(l)}}
        \exp\!\bigl(\boldsymbol{a}_i^{(l+1)} \cdot \boldsymbol{p} \,/\, \tau\bigr)
}{
    \displaystyle \sum_{\boldsymbol{c} \in \mathcal{P}_i^{(l)} \cup \mathcal{N}_i^{(l)}}
        \exp\!\bigl(\boldsymbol{a}_i^{(l+1)} \cdot \boldsymbol{c} \,/\, \tau\bigr)
}
\end{equation}

\noindent
Here, $\boldsymbol{p}$ is a positive sequence token from the anchor’s temporal window,  
$\boldsymbol{n} \in \mathcal{N}_i^{(l)}$ is a negative from distant anchors,  
$\cdot$ denotes cosine similarity, and $\tau$ is a temperature.  
Aggregating over all anchors and layers yields:
\begin{equation}
\label{eq:acc_layer}
\mathcal{L}_{\text{ACC}} = \sum_{l=0}^{L-1} \sum_{i} \mathcal{L}_{\text{acc}}(\boldsymbol{a}_i^{(l+1)}).
\end{equation}

\noindent
This contrastive signal propagates through the hierarchy, with ACC formulated for learning compact summary tokens designed for video localization. It aligns each anchor with all frames in its temporal window and uses negatives from other temporally distant anchors to enhance discriminability, producing embeddings that faithfully condense their window while remaining distinctive across events.

\paragraph{Segment-Pooled Contrastive (SPC) Loss.}
\label{sec:spc_loss}
While ACC provides unsupervised structural guidance, the Segment-Pooled Contrastive (SPC) loss uses ground-truth query spans to make representations of \textbf{ground-truth moments} discriminative against surrounding content. 
It encourages tokens within annotated segments to align, while pushing apart those from unrelated regions.

To achieve this, we construct a contrastive objective that, at each layer \(l\), considers every annotated segment
\(
g_m = [t_{\text{start}},\,t_{\text{end}})
\)
(the \(m\)-th ground-truth interval).
We aggregate the refined sequence tokens \(\tilde{\boldsymbol{v}}^{(l)}_t\) within this interval into a single segment prototype:
\[
\boldsymbol{z}_{\text{seg}}^{(l)} = 
\operatorname{Pool}\!\bigl\{\tilde{\boldsymbol{v}}^{(l)}_t \mid t \in g_m\bigr\}.
\vspace{-1mm}
\]
The prototype \(\boldsymbol{z}_{\text{seg}}^{(l)}\) serves as the \textbf{positive anchor}, contrasted against
\(
\mathcal{N}_{\text{seg}}^{(l)} = 
\{\tilde{\boldsymbol{v}}^{(l)}_t \mid t \notin g_m\}
\),
the set of tokens outside the ground-truth interval.
This formulation encourages tokens within the same segment to align through their shared prototype while repelling tokens from unrelated regions.

Importantly, using the pooled prototype as the positive target prevents diverse sub-motions within a segment (e.g., reaching, grasping, retracting) from collapsing into a single representation. 
Instead, it ties them to a common high-level event concept—a design choice crucial for preserving intra-segment diversity (see ablation in the supplementary).

Similar to ACC, all embeddings used in the loss are passed through a shared linear projection head.  
The objective for each segment at layer $l$ is:
\begin{equation}
\label{eq:spc_loss}
\mathcal{L}_{\mathrm{spc}}^{(l)}\!\bigl(\boldsymbol{z}_{\mathrm{seg}}^{(l)}\bigr) =
-\log \frac{
    \displaystyle \sum_{\boldsymbol{p} \in \mathcal{P}_{\mathrm{seg}}^{(l)}}
        \exp\!\bigl(\boldsymbol{z}_{\mathrm{seg}}^{(l)} \cdot \boldsymbol{p} \,/\, \tau\bigr)
}{
    \displaystyle \sum_{\boldsymbol{c} \in \mathcal{P}_{\mathrm{seg}}^{(l)} \cup \mathcal{N}_{\mathrm{seg}}^{(l)}}
        \exp\!\bigl(\boldsymbol{z}_{\mathrm{seg}}^{(l)} \cdot \boldsymbol{c} \,/\, \tau\bigr).
}
\end{equation}

Aggregating over all layers yields the SPC loss:
\begin{equation}
\label{eq:spc_total}
\mathcal{L}_{\text{SPC}} = \sum_{l=0}^{L-1} \mathcal{L}_{\text{spc}}^{(l)}.
\end{equation}

\paragraph{Putting it Together.} 
ACC provides layer-wise \emph{hierarchy consistency}: each anchor is pulled toward tokens in its window (compactness) and pushed from anchors of distant windows (distinctiveness).
SPC provides query-level \emph{semantic alignment}: segment prototypes from ground-truth spans are pulled toward in-segment tokens and pushed from surrounding context. We jointly minimize both objectives:
\begin{equation}
\label{eq:total_contrastive}
\mathcal{L}_{\text{contrast}} = \lambda_{\text{ACC}}\mathcal{L}_{\text{ACC}} \;+\; \lambda_{\text{SPC}}\mathcal{L}_{\text{SPC}},
\end{equation}

\noindent where $\lambda_{\text{ACC}}$ and $\lambda_{\text{SPC}}$ balance structural and semantic supervision.  
Together they yield anchors that are simultaneously \textbf{compact}, \textbf{distinctive}, and \textbf{query-aligned}, providing a robust foundation for downstream temporal grounding.

\begin{table*}[t!]
\centering
\small

\begin{minipage}[t]{0.34\linewidth}
\centering
\setlength{\tabcolsep}{4pt}
\resizebox{\linewidth}{!}{
\begin{tabular}{lccccc}
\toprule
\multirow{2}{*}{\textbf{Method}} 
& \multicolumn{2}{c}{\textbf{R@1}} 
& \multicolumn{2}{c}{\textbf{R@5}} 
& \multirow{2}{*}{\textbf{Avg.}} \\ 
& \textbf{0.3} & \textbf{0.5} & \textbf{0.3} & \textbf{0.5} & \\        
\midrule
2D-TAN~\cite{zhang2020learning} & 5.04 & 2.02 & 12.89 & 5.88 & 6.46 \\
VSLNet~\cite{zhang2020span} & 10.84 & 6.81 & 18.84 & 13.45 & 12.49 \\
M-DETR~\cite{lei2021detecting} & 8.23 & 5.01 & 23.23 & 13.37 & 12.46 \\
CONE~\cite{hou2022cone} & 14.15 & 8.18 & 30.33 & 18.02 & 17.67 \\
UniVTG~\cite{lin2023univtg} & 11.74 & 3.25 & 7.54 & 7.88 & 7.60 \\
SOONet~\cite{pan2023scanning} & 8.00 & 3.76 & 22.40 & 11.09 & 11.31 \\
H-Hands~\cite{zhang2023helping} & 13.20 & 7.90 & 23.30 & 15.60 & 15.00 \\
SnAG~\cite{snag} & 15.72 & 10.78 & 38.39 & 27.44 & 23.08 \\
RGNet~\cite{hannan2024rgnet} & 18.28 & 12.04 & 34.02 & 22.89 & 21.81 \\
DeCafNet~\cite{lu2025decafnet} & 18.10 & 12.55 & 38.85 & 28.27 & 24.44 \\
OSGNet~\cite{feng2025OSGNet} & 16.13 & 11.28 & 36.78 & 25.63 & 22.46 \\
\rowcolor{gray!12}
\textbf{HieraMamba (ours)} & \textbf{18.81} & \textbf{13.04} & \textbf{40.82} & \textbf{29.96} & \textbf{25.66} \\
\bottomrule
\end{tabular}
}
\caption{Results on \textbf{Ego4D-NLQ} using EgoVLP~\cite{lin2022egocentric} features.}
\label{tab:ego4d_nlq}
\end{minipage}\hfill
\begin{minipage}[t]{0.35\linewidth}
\centering
\setlength{\tabcolsep}{4pt}
\renewcommand{\arraystretch}{1.05}
\resizebox{\linewidth}{!}{
\begin{tabular}{clccccc}
\toprule
\multirow{2}{*}{\textbf{Version}} & \multirow{2}{*}{\textbf{Method}} 
& \multicolumn{2}{c}{\textbf{R@1}} 
& \multicolumn{2}{c}{\textbf{R@5}} 
& \multirow{2}{*}{\textbf{Avg.}} \\
& & \textbf{0.3} & \textbf{0.5} & \textbf{0.3} & \textbf{0.5} & \\
\midrule
\multirow{9}{*}{\textbf{MAD-v1}}
& 2D-TAN~\cite{zhang2020learning} & 2.52 & 1.58 & 9.25  & 5.69  & 4.76 \\
& VLG-Net~\cite{soldan2021vlg}    & 2.76 & 1.65 & 9.31  & 5.99  & 4.93 \\
& M-DETR~\cite{lei2021detecting}  & 2.81 & 1.67 & 9.86  & 5.58  & 4.98 \\
& CONE~\cite{hou2022cone}         & 6.87 & 4.10 & 16.11 & 9.59  & 9.17 \\
& SOONet~\cite{pan2023scanning}   & 9.00 & 5.32 & 19.64 & 3.14  & 9.28 \\
& SnAG~\cite{snag}                & 8.46 & 5.55 & 20.60 & 13.75 & 12.09 \\
& RGNet~\cite{hannan2024rgnet}    & 9.48 & 5.61 & 18.72 & 10.86 & 11.17 \\
& DeCafNet~\cite{lu2025decafnet}  & 10.96 & 7.06 & \textbf{23.68} & 16.13 & 14.46 \\
\rowcolor{gray!12}
& \textbf{HieraMamba (ours)}      & \textbf{11.26} & \textbf{7.22} & 23.49 & \textbf{16.81} & \textbf{14.70} \\
\midrule
\multirow{4}{*}{\textbf{MAD-v2}}
& CONE~\cite{hou2022cone}         & 9.70 & 5.43 & 20.31 & 11.41 & 11.71 \\
& SnAG~\cite{snag}                & 11.61 & 7.39 & 25.23 & 16.76 & 15.25 \\
& RGNet~\cite{hannan2024rgnet}    & 13.02 & 7.63 & 24.43 & 14.40 & 14.87 \\
\rowcolor{gray!12}
& \textbf{HieraMamba (ours)}      & \textbf{14.72} & \textbf{9.00} & \textbf{28.50} & \textbf{19.97} & \textbf{18.05} \\
\bottomrule
\end{tabular}
}
\caption{Results on \textbf{MAD} using CLIP~\cite{radford2021learning} features (ViT-B/32 for v1 following prior works, and ViT-L/14 for v2).}
\label{tab:mad}
\end{minipage}\hfill
\begin{minipage}[t]{0.30\linewidth}
\centering
\setlength{\tabcolsep}{4pt}
\resizebox{\linewidth}{!}{
\begin{tabular}{lccccc}
\toprule
\multirow{2}{*}{\textbf{Method}} 
& \multicolumn{2}{c}{\textbf{R@1}} 
& \multicolumn{2}{c}{\textbf{R@5}} 
& \multirow{2}{*}{\textbf{Avg.}} \\
& \textbf{0.3} & \textbf{0.5} & \textbf{0.3} & \textbf{0.5} & \\
\midrule
SMIN~\cite{wang2021structured} & 48.01 & 35.24 & 65.18 & 53.36 & 50.45 \\
CBLN~\cite{liu2021context} & 38.98 & 27.65 & 73.12 & 46.24 & 46.50 \\
MATN~\cite{zhang2021multi} & 48.79 & 37.57 & 67.63 & 57.91 & 52.98 \\
VLG-Net~\cite{soldan2021vlg} & 45.46 & 34.19 & 70.38 & 56.56 & 51.65 \\
APGN~\cite{liu2021adaptive} & 40.47 & 27.86 & 59.98 & 47.12 & 43.86 \\
IA-Net~\cite{liu2021progressively} & 37.91 & 26.27 & 57.62 & 46.39 & 42.05 \\
RaNet~\cite{gao2021relation} & 43.34 & 33.54 & 67.33 & 55.09 & 49.83 \\
MGSL-Net~\cite{liu2022memory} & 42.54 & 32.27 & 63.39 & 50.13 & 47.08 \\
MMN~\cite{wang2022negative} & 39.24 & 26.17 & 62.03 & 47.39 & 43.71 \\
SSRN~\cite{zhu2023rethinking} & 45.10 & 34.33 & 65.26 & 51.85 & 49.14 \\
G2L~\cite{li2023g2l} & 42.74 & 30.95 & 65.83 & 49.86 & 47.35 \\
MSAT~\cite{panta2024cross} & 49.77 & 37.99 & 68.31 & 58.31 & 53.60 \\
SnAG~\cite{snag} & 56.44 & 44.86 & 81.15 & 70.66 & 63.28 \\
DeCafNet~\cite{lu2025decafnet} & 57.36 & 46.79 & 81.05 & 71.13 & 64.11 \\
OSGNet~\cite{feng2025OSGNet} & 57.57 & 48.18 & 82.02 & 72.05 & 64.96 \\
\rowcolor{gray!12}
\textbf{HieraMamba (ours)} & \textbf{59.59} & \textbf{48.99} & \textbf{83.75} & \textbf{74.28} & \textbf{66.65} \\
\bottomrule
\end{tabular}
}
\caption{Results on \textbf{TACoS} using C3D~\cite{tran2015learning} features.}
\label{tab:tacos}
\end{minipage}

\vspace{-2mm}
\end{table*}

\section{Experimental Setup}\label{sec:results}
\vspace{-2mm}
We evaluate HieraMamba on diverse long-video grounding benchmarks and provide dataset, metric, efficiency, and ablation analyses for fair comparison.

\vspace{-1mm}
\subsection{Datasets}
\vspace{-1.5mm}
We evaluate our approach on three challenging long video temporal grounding benchmarks, 
Ego4D-NLQ~\cite{grauman2022ego4d}, MAD~\cite{soldan2022mad}, and TACoS~\cite{tacos}, which contain long videos with diverse queries that stress both scale and precision. 

\noindent\textbf{Ego4D‐NLQ}~\cite{grauman2022ego4d} is drawn from the large‐scale egocentric Ego4D corpus: it comprises videos recorded by 931 camera wearers in hundreds of daily scenarios, with clip lengths ranging from 3.5 to 20 minutes (avg.\ 8.3 min) and 74K natural‐language queries covering 13 question templates. Answer moments avg. 8.3s, only about $2\%$ of each video. 

\noindent\textbf{MAD}~\cite{soldan2022mad} comprises 488 full-length movies ($\approx$1.2K hours, avg.\ 110 min) and 384K timestamped audio-description queries. 
Its refined version, \textbf{MAD-v2}~\cite{han2023autoad}, reduces annotation noise to yield around 264K cleaner descriptions over the same movies (effective duration of 892 hours) and provides a 10-movie eval subset for clean evaluation. Answer moments in both average roughly 4s.

\noindent\textbf{TACoS}~\cite{tacos} comprises 10.1 hours of cooking videos across 127 clips (avg.\ 4.8 min) with 27K queries in the 10.1K/4.6K/4.1K train/val/test split ($\approx143.5$ queries per video). Answer moments average 27.9s.

\vspace{-1mm}
\subsection{Evaluation Metric and Implementation Details}
\vspace{-1.5mm}
\paragraph{Evaluation Metric.} Following prior work~\cite{anne2017localizing, soldan2022mad, grauman2022ego4d}, we evaluate grounding performance using the standard \textbf{Recall $k$ at IoU=$\theta$} metric. This metric computes the percentage of queries for which at least one of the top-$k$ predicted moments has a temporal intersection-over-union (tIoU) with the ground-truth moment exceeding a threshold $\theta$. 
Following \cite{snag, lu2025decafnet, hannan2024rgnet, hou2022cone, feng2025OSGNet}, we report Recall $k$@$\theta$ at $k \in \{1, 5\}$ and $\theta \in \{0.3, 0.5\}$ for all datasets. 

\vspace{-5mm}
\paragraph{Implementation Details.} We adopt the \emph{dataset-specific base features established in prior work for each benchmark}, ensuring consistency with standard practice and enabling direct comparison with existing results.
Specifically, for \textbf{Ego4D-NLQ} we use the video–text features from EgoVLP~\cite{lin2022egocentric}, extracted with a 32-frame window and a 16-frame stride from 30\,fps video~\cite{snag,hannan2024rgnet, lu2025decafnet, feng2025OSGNet}.
Models are trained on the official training split (without narration augmentations~\cite{ramakrishnan2023naq}) and evaluated on the validation split. 
For \textbf{MAD}, we adopt CLIP ViT-B/32 video features~\cite{radford2021learning} for v1 following prior work, and CLIP ViT-L/14 video features~\cite{radford2021learning} for v2. Both feature sets are publicly available and provided by~\cite{soldan2022mad}.
For \textit{MAD-v1}, we train on the official training split and evaluate on the test split. 
For \textit{MAD-v2}, we use the refined annotations from~\cite{han2023autoad}, which reduce noise in the original labels, and evaluate on the 10-movie \textit{eval} subset for clean comparison. 
Since no official MAD-v2 leaderboard exists, we independently reproduce all baseline results using each method’s released implementation and default MAD-v1 settings, including our own model. We include recent methods that provide end-to-end training and evaluation code~\cite{hou2022cone, snag, hannan2024rgnet}.
For \textbf{TACoS} we employ C3D video features~\cite{tran2015learning} and 300-d GloVe embeddings~\cite{pennington2014glove} for the queries.  
Video features are computed with a 16-frame window and a 16-frame stride at 30\,fps.
Across all datasets, we strictly follow the official evaluation protocols. 

\subsection{Comparison with State-of-the-Art}
\noindent\textbf{Ego4D-NLQ~\cite{grauman2022ego4d}.}
We present our main results on the Ego4D-NLQ validation set in Table~\ref{tab:ego4d_nlq}, following the standard protocol of training only on the official NLQ data for a fair comparison. 
HieraMamba establishes a new state-of-the-art, consistently outperforming prior methods. 
Notably, the gains are pronounced when compared to widely-used methods,
with our model outperforming SnAG~\cite{snag} by \textbf{2.58\%} and RGNet~\cite{hannan2024rgnet} by \textbf{3.85\%}. 
These gains are particularly meaningful on this ``needle-in-a-haystack'' benchmark where ground-truth moments occupy only \(\sim2\%\) of each video (\(\sim8.3\) s). A \(+\!1\)~pp gain in overall average recall corresponds to roughly \(N/100\) additional queries (\(N\!\approx\!3500\) on Ego4D-NLQ val) becoming correctly localized. 
This improvement reflects not only tighter boundary alignment but also new successful detections that prior methods miss.

\noindent\textbf{MAD~\cite{soldan2022mad}.}
HieraMamba achieves state-of-the-art performance on both the v1 and refined v2 splits (Table~\ref{tab:mad}), which comprise exceptionally long, hour-scale videos. On v2, it outperforms the strongest baseline, SnAG, by \textbf{+2.80\%} in average recall. These results highlight HieraMamba’s ability to preserve temporal fidelity while remaining computationally efficient even at extreme video durations.

\noindent\textbf{TACoS~\cite{tacos}.}
Our model's strong performance continues on the TACoS benchmark (Table~\ref{tab:tacos}), where HieraMamba outperforms all prior methods across every metric. It demonstrates the versatility of our hierarchical approach even on videos with highly complex and compositional actions.

\begin{figure}[H]
  \centering
  \includegraphics[width=0.99\linewidth]{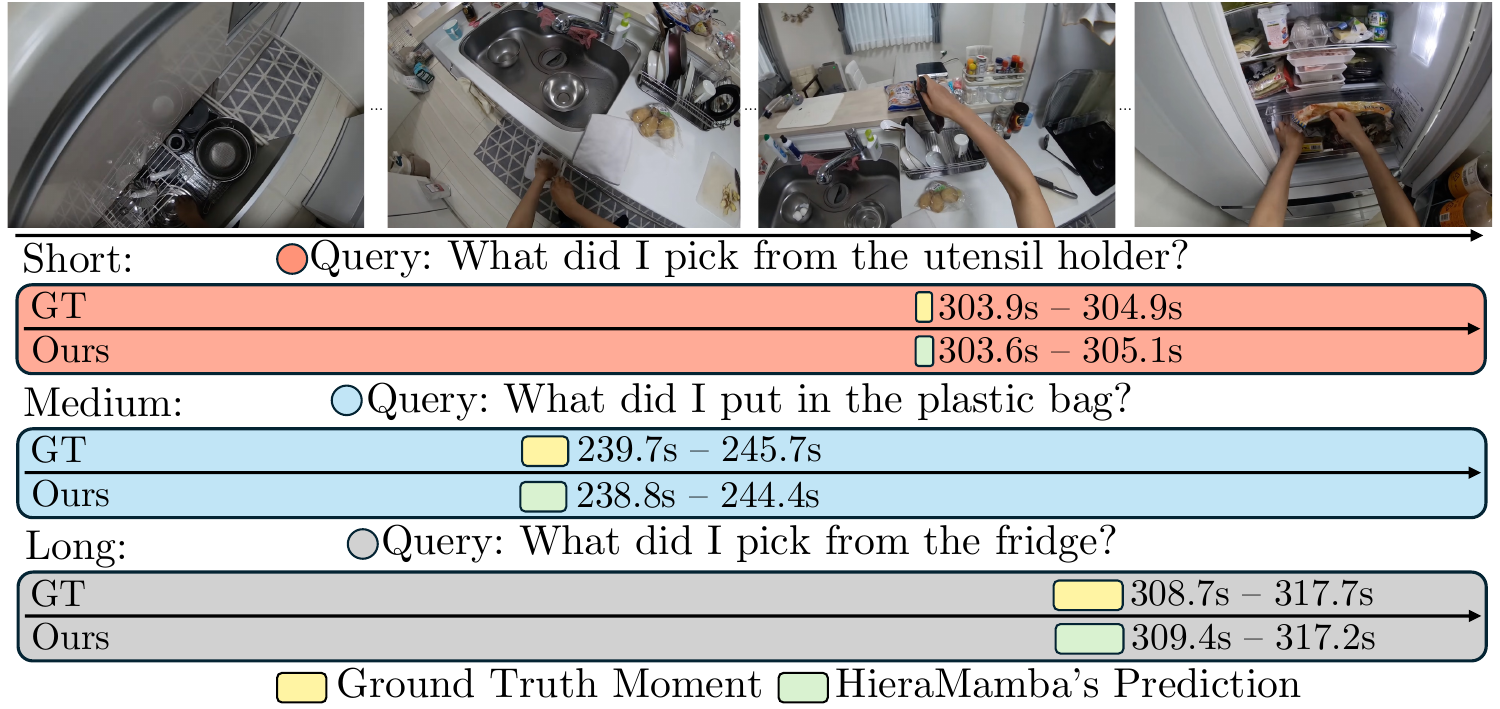}
  \vspace{-3mm}
    \caption{
        \textbf{Qualitative Visualization.} 
        Qualitative visualization of queries, ground truth, and predictions. Queries span diverse durations, and HieraMamba adapts flexibly across temporal scales; \textbf{additional examples and failure cases are shown in the supp}.
    }
  \label{fig:qualitative_visualize}
\end{figure}

\noindent\textbf{Summary of Results.}
Taken together, the consistent state-of-the-art performance across these three distinct benchmarks validates the robustness of HieraMamba. Its success on the sparse `needle-in-a-haystack' challenge of Ego4D-NLQ, the extreme duration of MAD, and the compositional complexity of TACoS demonstrates that our hierarchical architecture with its learned contrastive objectives is a powerful and generalizable approach for long-video temporal grounding.  See qualitative examples in Fig.~\ref{fig:qualitative_visualize} and supp.

\subsection{Efficiency and Scalability Analysis}
\label{sec:efficiency}
\vspace{-1mm}
To assess the practical viability of HieraMamba, we analyze its average recall versus computational cost (FLOPs) on \textbf{MAD-v2}~\cite{soldan2022mad, han2023autoad} eval, where videos average around 100 minutes, hence most straining complexity among all three datasets. We compare against the \textbf{strongest open-source baselines} RGNet~\cite{hannan2024rgnet} and SnAG~\cite{snag}. 
The default configuration of SnAG, which we denote \texttt{SnAG (Local)}, is restricted by a local self-attention window, handicapping its ability to model long-range dependencies. To build a stronger baseline, we modify its architecture to employ full, non-local self-attention, creating a \texttt{SnAG (Global)} variant that reasons over the entire video context.

\begin{figure}[t]
  \centering
  \includegraphics[width=0.9\linewidth]{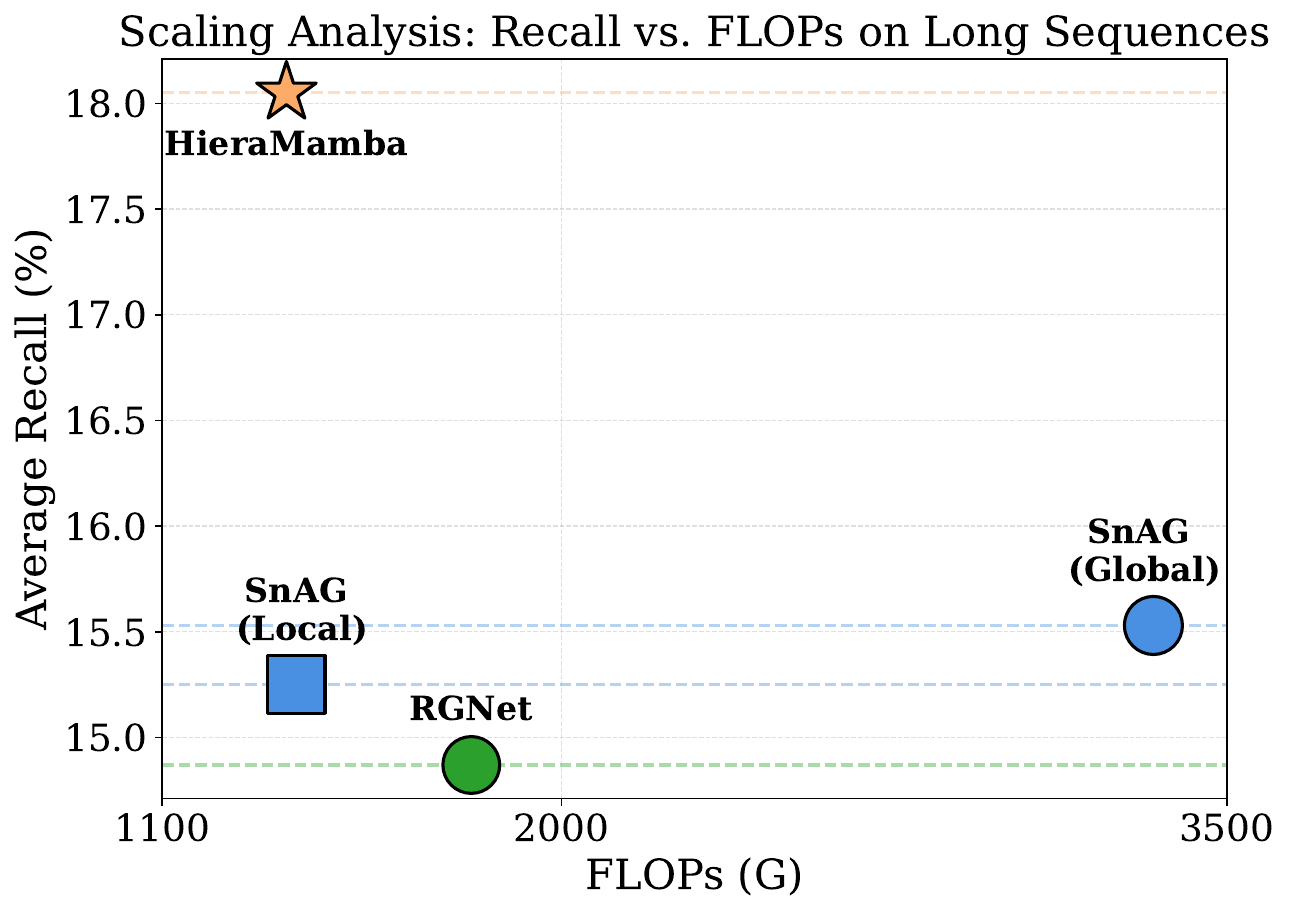}
  \vspace{-3mm}
    \caption{
        \textbf{Accuracy-Compute Trade-off.} 
        We plot average recall on the MAD-v2 eval set against computational cost (FLOPs) measured for a single forward pass on its average length video. HieraMamba achieves state-of-the-art accuracy with significantly lower computational cost than strong baselines.
    }
  \label{fig:compute_efficiency}
  \vspace{-1mm}
\end{figure}

The results in Figure~\ref{fig:compute_efficiency} highlight HieraMamba’s clear advantage. It achieves the highest accuracy while remaining efficient. Compared to its closest competitor, \texttt{SnAG (Global)}, HieraMamba improves average recall by +2.52\% while requiring roughly \textbf{2.5$\times$ fewer FLOPs}. Although the \texttt{SnAG (Global)} variant marginally outperforms \texttt{SnAG (Local)}, it does so at nearly a \textbf{3$\times$} increase in computational cost, underscoring the inefficiency of quadratic attention at this scale. In contrast, HieraMamba attains superior accuracy while processing the full video context in a far more efficient, linear-time manner.

This analysis shows that HieraMamba not only achieves a better trade-off but also pushes the Pareto frontier of accuracy and efficiency. Its Anchor-MambaPooling design exploits Mamba’s linear-time dynamics to capture long-range dependencies without the prohibitive cost of self-attention, making it both accurate and scalable.
These gains are best understood relative to existing long-video grounding baselines, RGNet~\cite{hannan2024rgnet} and SnAG~\cite{snag}, the strongest open-source baselines for long, untrimmed videos. While Mamba-based encoders like VideoMamba~\cite{li2024videomamba} also use linear-time sequence modeling, HieraMamba uniquely applies this efficiency to query-conditioned grounding, isolating its hierarchical design as the source of advantage.

\subsection{Ablation Studies}
We conduct ablation studies to validate HieraMamba’s key design choices and quantify the contribution of its core components. Results in the main tables are reported on Ego4D-NLQ, which offers the most diverse queries and video durations; consistent trends are observed on other datasets, presented in the supp.

We begin by demonstrating the superiority of HieraMamba over naive Mamba adaptations. Table~\ref{tab:ssm_baselines} reports two SSM-based baselines to show that our gains go beyond merely applying Mamba: (i) \emph{VideoMamba+SnAG}, which adapts the general-purpose VideoMamba~\cite{li2024videomamba} for VTG by equipping it with a SnAG~\cite{snag} grounding model, and (ii) \emph{SnAG-Mamba}, which replaces SnAG's Transformers with Hydra~\cite{hydra}, the SSM variant adopted in HieraMamba. Strategy (i) degrades performance, and crucially, strategy (ii) yields \emph{no gain}. This validates that standard SSMs alone cannot solve the VTG bottleneck. HieraMamba's gains stem specifically from AMP's hierarchical compression, highlighting a structural advantage that naive Mamba baselines cannot achieve.

\begin{table}[t]
  \centering
  
  \begin{minipage}[t]{0.41\linewidth}
    \centering
    \resizebox{\linewidth}{!}{%
    \begin{tabular}{cc}
      \toprule
      \textbf{Method} & \textbf{Avg.} \\
      \midrule
      SnAG & 23.14 \\
      \begin{tabular}[c]{@{}c@{}}VideoMamba+SnAG\end{tabular} & 12.44 \\
      SnAG-Mamba & 22.88 \\
      \midrule
      HieraMamba & \textbf{25.66} \\
      \bottomrule
    \end{tabular}%
    }
    \vspace{-2.5mm}
    \caption{\textbf{Mamba baselines.} Results of naively adapting Mamba to VTG.}
    \label{tab:ssm_baselines}
  \end{minipage}\hfill
  \begin{minipage}[t]{0.55\linewidth}
    \centering
    \setlength{\tabcolsep}{4pt}
    \resizebox{\linewidth}{!}{%
    \begin{tabular}{l S[table-format=2.2] S[table-format=+2.2]}
      \toprule
      \textbf{Variant}                     & {\textbf{Avg.}} & {\textbf{$\Delta$}} \\
      \midrule
      HieraMamba (Full)                    & 25.66  & {}        \\
      \addlinespace[0.3em]
      \hspace{0.6em}w/o Interleaving       & 24.40  & -1.26     \\
      \hspace{0.6em}w/o Bidirectional Scan & 23.29  & -2.37     \\   
      \hspace{0.6em}w/o Local Encoding     & 24.63  & -1.03     \\   
      \hspace{0.6em}w/o Gates              & 24.80  & -0.86     \\
      \bottomrule
    \end{tabular}%
    }
    \vspace{-2.5mm}
    \caption{\textbf{Ablation Study.} 
             We remove one component at a time.}
    \label{tab:amp_ablation}
  \end{minipage}
  
  \vspace{-1mm}
\end{table}

\begin{table}[t]
  \centering
  
  \begin{minipage}[t]{0.26\linewidth}
    \centering
    \small
    \setlength{\tabcolsep}{3pt}
    \resizebox{\linewidth}{!}{%
    \begin{tabular}{ccc} 
      \toprule
      ACC & SPC & \textbf{Avg.} \\
      \midrule
      $\times$ & $\times$ & 24.68 \\   
      \checkmark & $\times$ & 25.22 \\
      $\times$ & \checkmark & 25.23 \\
      \rowcolor{gray!15}
      \checkmark & \checkmark & \textbf{25.66} \\
      \bottomrule
    \end{tabular}%
    }
    \vspace{-3mm}
    \caption{\textbf{Contrastive Losses Ablation.}}
    \label{tab:ablation_losses}
  \end{minipage}\hfill
  \begin{minipage}[t]{0.39\linewidth}
    \centering
    \resizebox{\linewidth}{!}{%
    \begin{tabular}{cc|cc}
      \toprule
      $\lambda_{ACC}$ & \textbf{Avg} & $\lambda_{SPC}$ & \textbf{Avg} \\
      \midrule
      0.1 & 24.68 & 0.01 & 24.75 \\
      1   & 24.89 & 0.1  & 24.82 \\
      \textbf{10} & \textbf{25.22} & \textbf{1} & \textbf{25.23} \\
      50  & 24.76 & 5    & 24.85 \\
      100 & 24.66 & 10   & 24.54 \\
      \bottomrule
    \end{tabular}%
    }
    \vspace{-3mm}
    \caption{\textbf{Coefficient Sweeps.} Results of varying $\lambda_{ACC \& SPC}$.} 
    \label{tab:loss_sweep}
  \end{minipage}\hfill
  \begin{minipage}[t]{0.33\linewidth}
    \centering
    \resizebox{\linewidth}{!}{%
    \begin{tabular}{ccc}
      \toprule
      \textbf{\# L} & \textbf{Avg} & \textbf{Params (M)} \\
      \midrule
      2  & 21.80 & 8.05 \\
      4  & 23.92 & 15.11 \\
      \textbf{8}  & \textbf{25.66} & 29.24 \\
      10 & 24.84 & 36.31 \\
      \bottomrule
    \end{tabular}%
    }
    \vspace{-3mm}
    \caption{\textbf{Depth sweep.} Results of varying AMP layers.}
    \label{tab:amp_depth}
  \end{minipage}
  
  \vspace{-5mm}
\end{table}

Table~\ref{tab:amp_ablation} assesses the contribution of each AMP component, showing that removing any element degrades performance, confirming their complementary roles. Table~\ref{tab:ablation_losses} further examines the effect of the two contrastive objectives, finding that both improve performance and their combination yields the best overall results, with their optimal coefficients determined via independent parameter sweep (Table~\ref{tab:loss_sweep}). Finally, Table~\ref{tab:amp_depth} sweeps the number of AMP layers to assess performance and parameter trade-offs.

\vspace{-2mm}
\section{Conclusion}
\vspace{-2mm}
We introduced HieraMamba, a linear-time architecture for long-video temporal grounding that preserves temporal fidelity and remains scalable.
With hierarchical Anchor-MambaPooling blocks and anchor-conditioned and segment-pooled contrastive losses, our model learns compact, semantically rich representations across temporal scales.
Experiments on Ego4D-NLQ, MAD, and TACoS show that HieraMamba surpasses prior state-of-the-art methods in both accuracy and efficiency.

Beyond temporal grounding, the AMP block offers a general framework for hierarchical, context-aware representation learning and is extendable to other video understanding tasks requiring reasoning across temporal scales.
Future directions include adaptive anchor generation that dynamically allocates temporal resolution based on content.

\section*{Acknowledgements} 
This research was supported in part by the UT Austin IFML NSF AI Institute.

{
    \small
    \bibliographystyle{ieeenat_fullname}
    \bibliography{main}
}

\clearpage
\setcounter{page}{1}
\maketitlesupplementary
\setcounter{section}{0}
\section{Additional Ablation Studies}
We present additional ablation studies to isolate and assess the design choices of our proposed modules and losses. Unless otherwise specified, all experiments are conducted on Ego4D-NLQ using the base HieraMamba architecture (without auxiliary losses) to isolate component effects, with results averaged over five runs.

\begin{table}[b]
\centering
\resizebox{\columnwidth}{!}{
\begin{tabular}{lcccccc}
\toprule
\multirow{2}{*}{Pooling Method} & \multicolumn{2}{c}{R@1} & \multicolumn{2}{c}{R@5} & \multicolumn{1}{c}{Average} \\
& 0.30 & 0.50 & 0.30 & 0.50 & R@1\&5 \\
\midrule
Mean Pooling       & 18.23 & 12.55 & 39.13 & 28.78 & 24.68 \\
Max Pooling        & 17.87 & 12.66 & 39.09 & 29.00 & 24.65 \\
Attention Pooling  & 17.63 & 12.28 & 38.93 & 29.00 & 24.46 \\
Gated Pooling      & 17.41 & 12.36 & 39.04 & 28.65 & 24.37 \\
\bottomrule
\end{tabular}
}
\caption{Comparison of pooling methods on retrieval performance (R@1, R@5, and average of R@1 \& R@5).}

\label{tab:pooling_methods}
\end{table}

\subsection{Anchor Generation Strategies}
\label{sec:anchor-gen}
We evaluate four strategies for generating anchors within the AMP block. 
Given a temporal stride $s$, each anchor is computed from its corresponding $s$ input tokens using one of the following pooling methods:  
(1)~\emph{Mean pooling}, which averages token features;  
(2)~\emph{Max pooling}, which selects the maximum activation per channel;  
(3)~\emph{Attention pooling}, which applies multi-head attention with a learnable query vector, following the attention pooling in CLIP~\cite{radford2021learning}; and  
(4)~\emph{Gated pooling}, which adaptively blends mean- and max-pooled features via a learned gate.

Table~\ref{tab:pooling_methods} reports the performance of the base HieraMamba model when applying each pooling strategy to its AMP blocks. 
For a fair comparison, no additional ACC or SPC losses are applied, isolating the effect of the pooling strategy itself.

Interestingly, the best results are obtained with non-learned pooling methods (mean and max), with mean pooling slightly outperforming max pooling. 
In contrast, learned variants (attention and gated pooling) underperform, with attention pooling yielding marginally better results than gated pooling. 
This suggests that simple statistical aggregation produces more stable anchors by avoiding early information loss, allowing the AMP’s temporal modeling blocks (global and local encoders) to compress and extract the most salient content.

\subsection{Impact of Pooling in Segment-Pooled Contrastive Loss}
\label{sec:spc_pooling_ablation}

To assess the role of pooling in our Segment-Pooled Contrastive (SPC) loss, we compare the proposed pooled formulation (\S\ref{sec:spc_loss}) 
with an \emph{unpooled} variant. In the unpooled setup, rather than contrasting the pooled segment prototype $\boldsymbol{z}_{\mathrm{seg}}^{(l)}$ against all tokens in the ground-truth interval, we treat every in-segment token as an independent positive example. This removes the aggregation step, effectively forcing all tokens within the same ground-truth moment to be pulled tightly together in the embedding space.

Table~\ref{tab:spc_variants} shows that the unpooled variant underperforms the pooled one, and even degrades the base model’s performance (HieraMamba without SPC or ACC losses). We attribute this drop to the fact that tokens within a ground-truth interval often correspond to distinct sub-actions (e.g., reaching, grasping, retracting) that should retain some temporal diversity. Forcing these heterogeneous sub-motions to collapse into a single point can blur fine-grained temporal dynamics, harming retrieval accuracy. 

By contrast, our pooled formulation produces a holistic, high-level segment representation, which is then contrasted against positives and negatives at the segment level. This design preserves intra-moment variability while still providing strong query-level semantic guidance, encouraging ground-truth moments to be discriminative to surrounding, non-matching content.

\begin{table}[H]
\centering
\resizebox{\columnwidth}{!}{
\begin{tabular}{lcccccc}
\toprule
\multirow{2}{*}{Method} & \multicolumn{2}{c}{R@1} & \multicolumn{2}{c}{R@5} & \multicolumn{1}{c}{Average} \\
& 0.30 & 0.50 & 0.30 & 0.50 & R@1\&5 \\
\midrule
HieraMamba (base) & 18.23 & 12.55 & 39.13 & 28.78 & 24.68 \\
+ SPC Loss (Pooled) & \textbf{18.52} & \textbf{13.01} & \textbf{39.99} & \textbf{29.39} & \textbf{25.23} \\
\rowcolor{gray!10} 
+ SPC Loss (UnPooled) & 17.23 & 11.77 & 38.95 & 28.24 & 24.05 \\
\bottomrule
\end{tabular}
}
\caption{Comparison of SPC loss variants on retrieval performance. }
\label{tab:spc_variants}
\end{table}

\begin{table}[t]
\centering
\small
\setlength{\tabcolsep}{3pt}
\resizebox{\columnwidth}{!}{%
\begin{tabular}{cccccccc}
\toprule
\textbf{Dataset} &
\multicolumn{2}{c}{\textbf{Components}} &
\multicolumn{4}{c}{\textbf{Recall (\%)}\,$\uparrow$} &
\textbf{Avg.} \\
\cmidrule(lr){2-3} \cmidrule(lr){4-7}
& ACC & SPC & R1@0.30 & R1@0.50 & R5@0.30 & R5@0.50 & \\ 
\midrule

\multirow{4}{*}{Ego4D}
& $\times$ & $\times$ & 18.23 & 12.55 & 39.13 & 28.78 & 24.68 \\
& \checkmark & $\times$ & 18.52 & 13.24 & 39.62 & 29.50 & 25.22 \\
& $\times$ & \checkmark & 18.52 & 13.01 & 39.99 & 29.39 & 25.23 \\
\rowcolor{gray!15}
& \checkmark & \checkmark & 18.81 & 13.04 & 40.82 & 29.96 & 25.66 \\
\midrule

\multirow{4}{*}{MADv2}
& $\times$ & $\times$ & 14.36 & 8.70 & 27.38 & 18.73 & 17.29 \\
& \checkmark & $\times$ & 14.82 & 9.19 & 28.17 & 19.43 & 17.90 \\
& $\times$ & \checkmark & 14.86 & 8.63 & 27.32 & 18.60 & 17.35 \\
\rowcolor{gray!15}
& \checkmark & \checkmark & 14.72 & 9.00 & 28.50 & 19.97 & 18.05 \\
\midrule

\multirow{4}{*}{TACoS}
& $\times$ & $\times$ & 58.29 & 49.16 & 83.10 & 72.08 & 65.66 \\
& \checkmark & $\times$ & 58.84 & 48.26 & 83.90 & 73.88 & 66.22 \\
& $\times$ & \checkmark & 58.99 & 48.64 & 82.78 & 73.21 & 65.81 \\
\rowcolor{gray!15}
& \checkmark & \checkmark & 59.59 & 48.99 & 83.75 & 74.28 & 66.65 \\
\bottomrule
\end{tabular}}
\caption{\textbf{Ablation of contrastive objectives across datasets.}}
\label{tab:ablation_losses2}
\end{table}

\subsection{Effect of ACC and SPC on Additional Datasets}
We evaluate the contribution of the two contrastive losses across all datasets. Table~\ref{tab:ablation_losses2} shows that each loss independently improves performance, and combining them achieves the strongest overall results.

\section{Additional Implementation Details}
We provide additional implementation details omitted from the main paper due to space constraints. Complete configurations and code are available in our official release.

\subsection{AMP Details}
As described in the main paper, we use Hydra~\cite{hydra} as the global encoder and a windowed Transformer~\cite{zhang2022actionformer} as the local encoder. For Hydra, we set $d_{\text{state}}=64$, $d_{\text{conv}}=7$, $\text{expand}=2$, and $\text{head\_dim}=64$. For the local encoder, we configure a single layer ($\text{num\_layers}=1$) with a small attention window ($\text{window\_size}=5$), $n_{\text{heads}}=2$, and $\text{stride}=1$, enabling it to focus on very local context while remaining lightweight due to its minimal window size, head dimension, and depth. We stack these AMP blocks to construct the multi-scale video pyramid (Multi-Scale Video Encoder in Fig.~\ref{fig:main}, left), using 8, 8, and 9 layers for Ego4D, TACoS, and MAD, respectively.

\subsection{Training Details}
We adopt the same training and inference settings as prior work~\cite{snag}, including learning rate, number of epochs, and other hyperparameters. Below, we detail the moment decoding procedure and the loss functions used for optimization.
\\

\noindent\textbf{Moment decoding.}
At each scale $l$, the refined sequence $\tilde V^{(l)}=\{\tilde{\boldsymbol{v}}^{(l)}_t\}_{t=1}^{L_l}$ is passed through two lightweight heads (three 1D convolutions each):  
(i) a classification head that outputs a confidence score $p_t^{(l)}$, and  
(ii) a regression head that predicts normalized start/end offsets $\boldsymbol{\delta}_t^{(l)} = (\delta^s, \delta^e)$.  
For brevity, we omit $(t,l)$ when clear from context.

Given the effective stride $S^{(l)}$ (e.g., $S^{(l)} = s^{\,l-1}$ for geometric downsampling by $s$), each token produces a proposal  
\[
\hat{\mathbf{y}} = \bigl(S^{(l)}(t - \delta^s),\; S^{(l)}(t + \delta^e)\bigr).
\]
We rank all proposals across $t$ and $l$ by $p$, and apply Soft-NMS~\cite{bodla2017softnms} over the multi-scale set to merge overlapping candidates, following common practice in video grounding~\cite{zhang2022actionformer,snag}.  
The final output consists of the top-$k$ moment predictions $\{(t_s, t_e)\}_{k=1}^K$ after Soft-NMS re-ranking.
\\

\noindent\textbf{Training objectives.}
The model is optimized with three loss terms:  
(i) a classification loss $\mathcal{L}_{\text{cls}}$ using Focal Loss~\cite{lin2017focal},  
(ii) a regression loss $\mathcal{L}_{\text{reg}}$ using Distant IoU Loss~\cite{zheng2020distance}, and  
(iii) a contrastive loss $\mathcal{L}_{\text{contrast}}$ that combines the proposed ACC and SPC losses.
$\mathcal{L}_{\text{contrast}}$ is as defined in Eq.~\ref{eq:total_contrastive} of the main paper, which are controlled by $\lambda_{\text{ACC}}$ and $\lambda_{\text{SPC}}$.
We set $(\lambda_{\text{ACC}}, \lambda_{\text{SPC}})$ to $(10, 1)$ for Ego4D, $(1, 0.1)$ for TACoS, and $(0.5, 0.6)$ for MAD.  
The final training objective is
\[
\mathcal{L} = \mathcal{L}_{\text{cls}} + \mathcal{L}_{\text{reg}} + \mathcal{L}_{\text{contrast}}.
\]

\section{Qualitative Results.}
In this section, we present qualitative visualizations of our model’s predictions for diverse language queries across a variety of scenarios. We use the Ego4D-NLQ~\cite{grauman2022ego4d} benchmark, where the ground-truth moment length can range from as short as one second to over 30 seconds, depending on the query and scenario. We first compare our visualizations against those from SnAG~\cite{snag}, the state-of-the-art open-source model for which we can run experiments, then provide additional visualizations showcasing our own predicted moments.

\begin{figure*}[!t]
  \centering
  \includegraphics[width=0.98\linewidth]{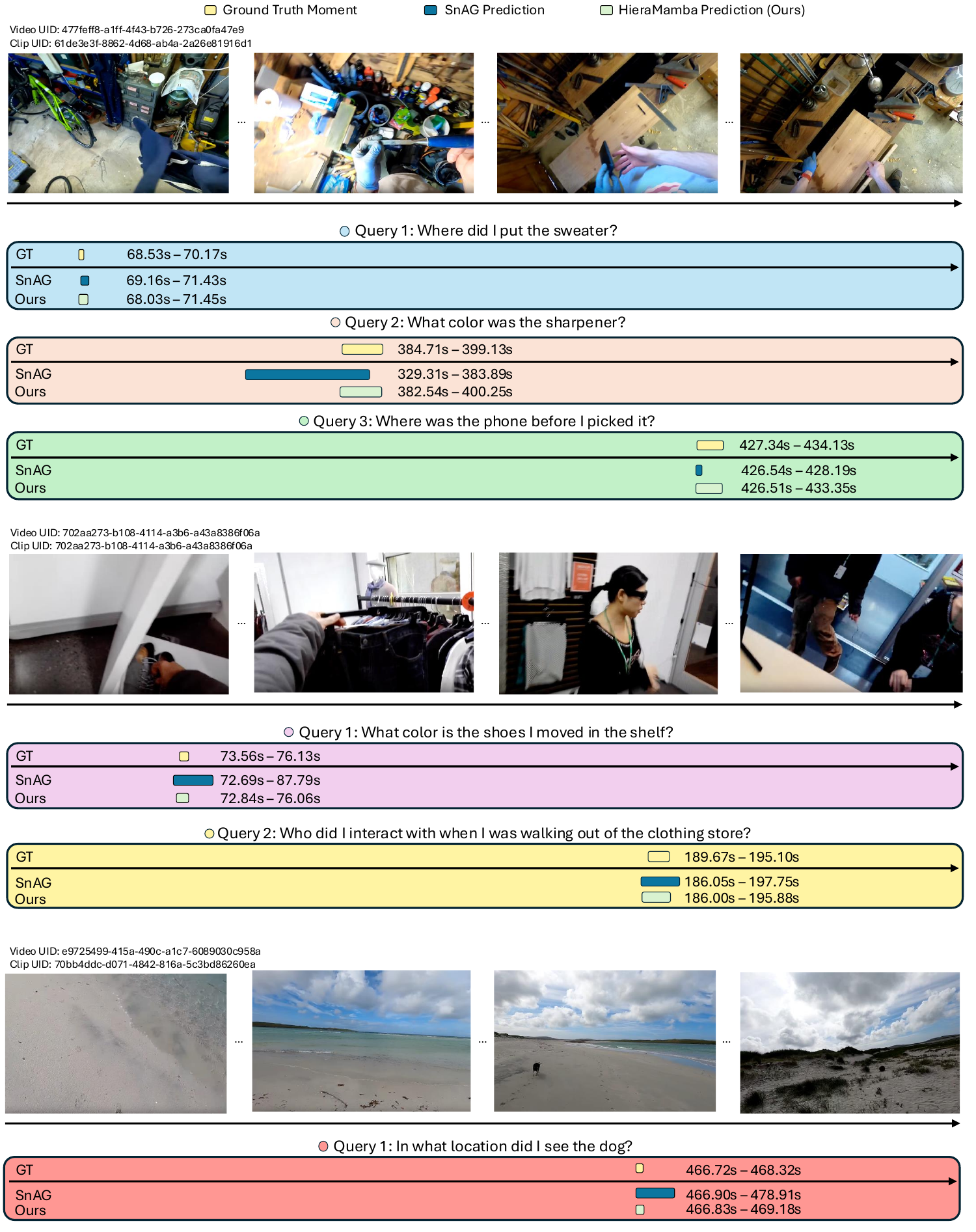}
    \caption{
        \textbf{Qualitative Results Comparison with SnAG~\cite{snag}.} 
    }
  \label{fig:qualitative_comparison}
\end{figure*}

\subsection{Qualitative Comparison with State-of-the-Art}
Figure~\ref{fig:qualitative_comparison} presents a side-by-side qualitative comparison between SnAG~\cite{snag} and our HieraMamba model. Each colored bar corresponds to a different language query for a given video clip: the yellow segment marks the ground-truth moment, the blue segment (beneath the yellow) shows SnAG’s prediction, and the green segment (final row) depicts our prediction.

The examples span diverse scenarios from the Ego4D-NLQ benchmark, where ground-truth moments range from fleeting events lasting barely a second to extended activities exceeding 30 seconds. This diversity demands a model capable of reasoning over both fine-grained and long-range temporal contexts. By leveraging hierarchical semantic representations across multiple temporal scales, our model effectively adapts to this variability—capturing the precise span for short events while maintaining coherence for extended activities.

In many cases, SnAG’s predictions exhibit partial misalignment with the ground truth, starting too early, ending prematurely, or drifting away from the relevant content. In contrast, HieraMamba’s predictions remain closely aligned with the annotated intervals across all temporal ranges. For example, in Query 2 of the first clip, SnAG localizes the moment too early, omitting critical visual evidence, whereas our method covers the complete span. Similarly, in the clothing store example, our prediction preserves the full interaction interval, avoiding the truncation seen in SnAG’s output. Even in cases where both predictions are close to the ground truth (e.g., second query in the clothing store scenario), our boundaries are slightly more precise, reflecting improved temporal alignment.

Overall, these qualitative results illustrate how multi-scale temporal reasoning enables HieraMamba to robustly localize events of vastly different durations, providing faithful and semantically coherent grounding across a wide variety of queries and scenarios.

\begin{figure*}[!t]
  \centering
  \includegraphics[width=0.87\linewidth]{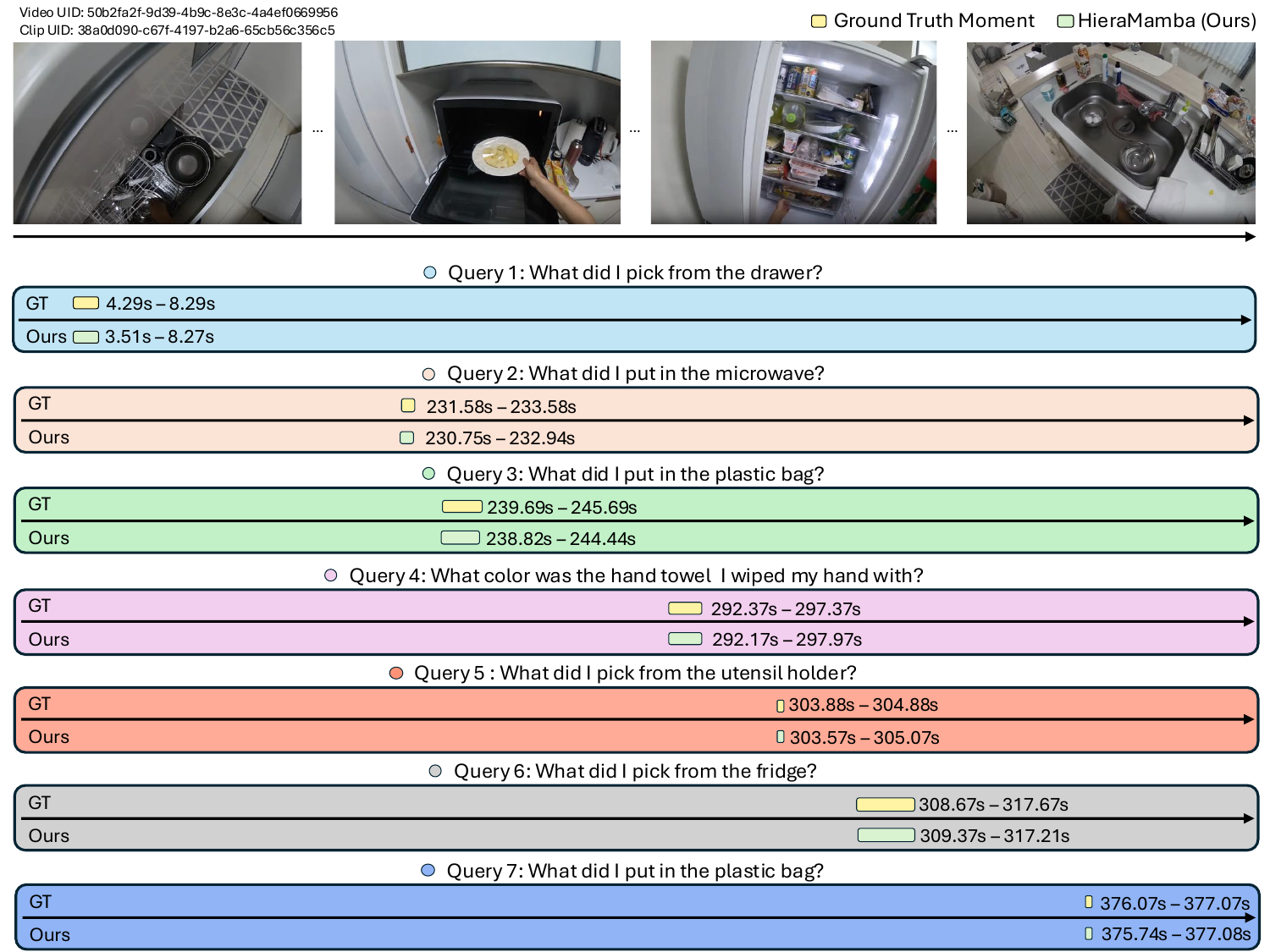}
    \caption{Qualitative Results 
    }
  \label{fig:qualitative_results1}
\end{figure*}

\begin{figure*}[!t]
  \centering
  \includegraphics[width=0.87\linewidth]{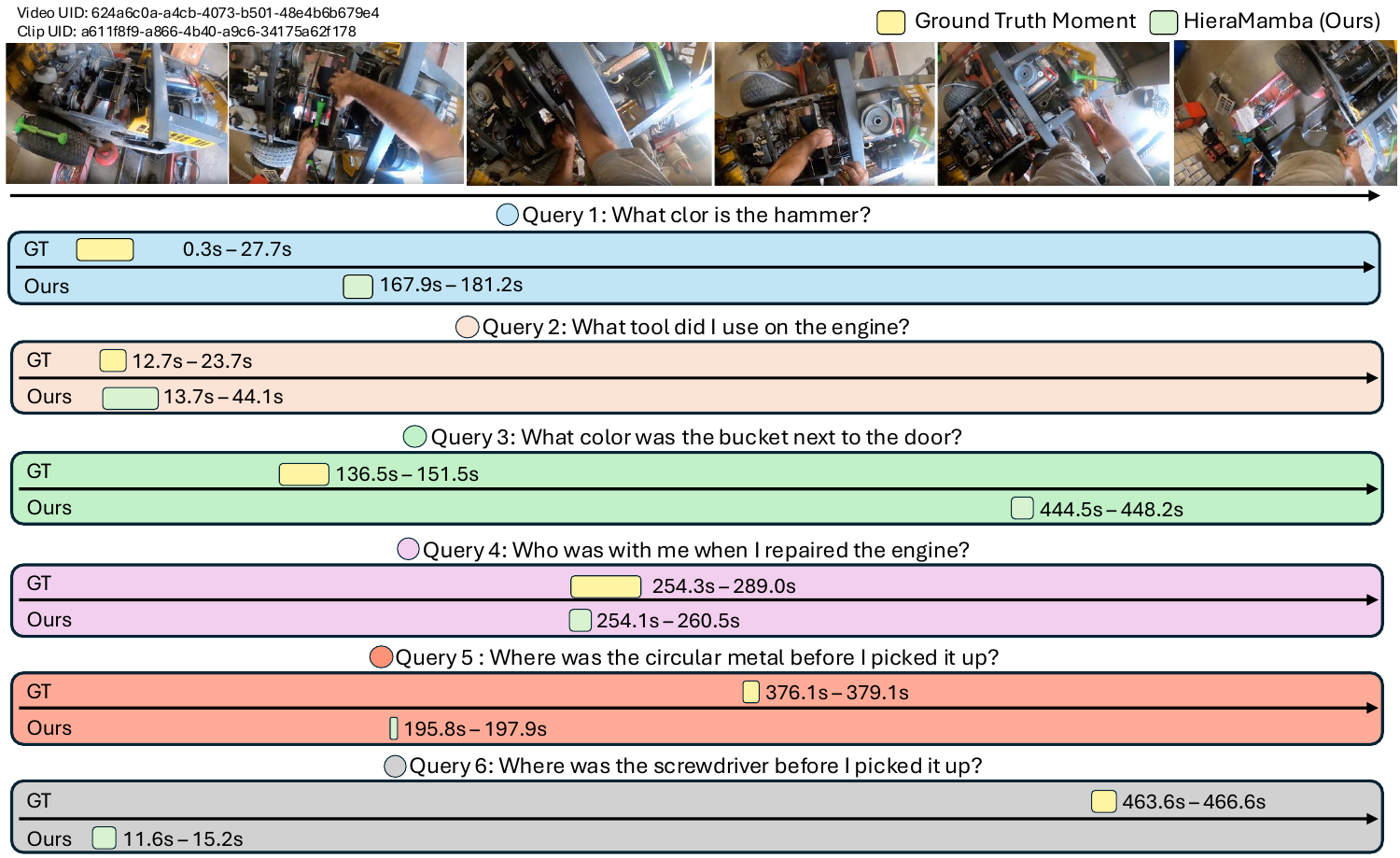}
    \caption{Qualitative Failure Examples.
    }
  \label{fig:qualitative_failures}
\end{figure*}

\begin{figure*}[!t]
  \centering
  \includegraphics[width=1\linewidth]{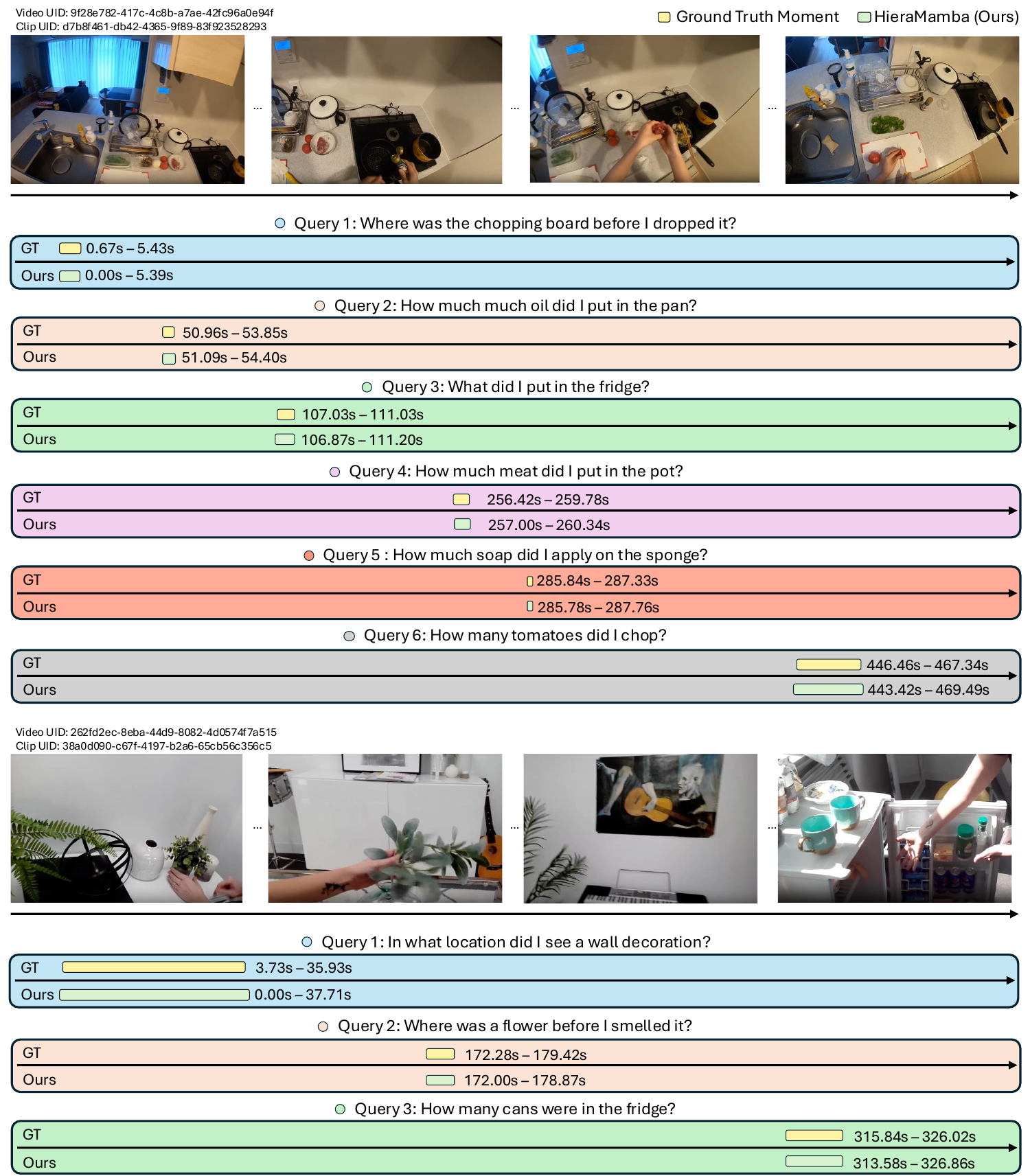}
    \caption{More qualitative results.
    }
  \label{fig:qualitative_results2}
\end{figure*}

\subsection{Qualitative Results: Handling Diverse Temporal Granularities}
Figures~\ref{fig:qualitative_results1} and~\ref{fig:qualitative_results2} show qualitative examples from Ego4D-NLQ demonstrating our model’s ability to localize moments of vastly different durations, even within the same continuous video. In realistic egocentric recordings, multiple queries can refer to events at very different temporal scales: a brief action lasting about a second (e.g., picking up an item) may appear alongside an extended activity exceeding 30 seconds (e.g., a multi-step cooking or interaction sequence). This variation arises not only across different videos, but also frequently within the same video, making accurate localization particularly challenging.

HieraMamba addresses this challenge by producing semantically rich representations at multiple temporal scales—capturing fine-grained details for short moments while also maintaining coherent long-range context for extended activities. This multi-scale representation enables the model to adapt its grounding behavior based on the temporal demands of each query, without sacrificing precision for short events or coverage for long events.

As shown in the figures, our predictions align closely with the ground truth across a wide range of temporal granularities. For short-duration queries, boundaries are tightly matched to the relevant frames; for long-duration queries, the predicted segments span the full relevant context without truncation or drift. These highlight our model’s ability to seamlessly navigate between fine and coarse temporal reasoning, a capability essential for handling the mixed temporal demands present in real-world scenarios.

\subsection{Qualitative Results: Failure Cases}
Figures~\ref{fig:qualitative_failures} show qualitative failure examples from Ego4D-NLQ~\cite{grauman2022ego4d}. For the first query, ``What color is the hammer?'', both intervals contain the same hammer, but the predicted moment corresponds to the hammer being actively used rather than stationary. The model appears biased toward dynamic usage cues instead of static appearances.

For the second query, ``What tool did I use on the engine?'', the model predicts a much longer interval because the tool and engine remain in view well beyond the actual interaction. This suggests that the model conflates object visibility with tool usage.

For ``What color was the bucket next to the door?'', the ground-truth bucket is upside-down and partially visible at the frame edge, requiring implicit 3D reasoning. The predicted segment instead shows another bucket in a clearer, more canonical view, which the model favors over the subtler ground-truth configuration.

For the interpersonal query ``Who was with me when I repaired the engine?'', the model truncates the interval when the other person becomes temporarily occluded due to camera motion, revealing a lack of persistent 3D environment understanding.

For ``Where was the circular metal before I picked it up?'', the model predicts a moment involving a small idler pulley, likely confusing it with the referenced circular metal due to similar shape and cluttered tool context.

Finally, for ``Where was the screwdriver before I picked it up?'', the model focuses on a moment where a screwdriver is visible near a dead-blow hammer being lifted. The similar shape and nearby motion lead the model to incorrectly associate the action with the screwdriver.

These failure cases highlight several systematic limitations. The model often conflates visibility with action boundaries, struggles with fine-grained object discrimination, and treats temporarily occluded objects or people as absent. It also favors dynamic tool usage over static attributes and has difficulty interpreting non-canonical viewpoints that require implicit 3D reasoning. Together, these patterns suggest that achieving reliable moment localization will require better modeling of action semantics, object identity, occlusion, and spatial geometry.

\section{Limitations}\label{sec:limits}

While HieraMamba provides a scalable and accurate framework for long-video temporal grounding, it also has several limitations that open avenues for future work. First, although our model achieves linear-time complexity and supports multi-scale reasoning, it relies on frozen video backbones. This modular design offers flexibility in selecting video encoders but also decouples video feature learning from the temporal grounding objective. Jointly fine-tuning the video backbone together with our model could further improve performance, though at the expense of the substantial compute required for training large backbone models.

Second, our anchor generation strategy operates with a fixed temporal stride. An adaptive mechanism that adjusts the stride dynamically based on video content, allocating more anchors to regions with higher temporal density and fewer to less informative segments, could further enhance localization accuracy and efficiency.
\clearpage

\end{document}